\pdfoutput=1
\documentclass[11pt]{article}
\usepackage[final]{acl}
\usepackage{times}
\usepackage{latexsym}
\usepackage[T1]{fontenc}
\usepackage[utf8]{inputenc}
\usepackage{microtype}
\usepackage{inconsolata}
\usepackage{booktabs}
\usepackage{multirow}
\usepackage{graphicx}
\usepackage{amsmath}
\usepackage{xcolor}
\usepackage{enumitem}
\usepackage{tikz}
\usetikzlibrary{positioning,arrows.meta,fit,backgrounds}
\usepackage{pgfplots}
\pgfplotsset{compat=1.17}
\usepgfplotslibrary{groupplots}
\usepackage{amssymb}
\setlist{nosep,leftmargin=*}

\newcommand{\sys}{KnowForge}
\newif\iflong
\longtrue   % arXiv bundle: extended version
\iflong\fi   % (extended build) avoid stretched vertical gaps mid-column next to full-width floats
\hypersetup{pdftitle={Locate, Judge, Read: Role-Specialized Tool Agents over a Knowledge-Point Index for Open-Domain Multimodal Document QA},
  pdfauthor={Wenlin Lin, Fanzhe Wei}}

\title{Locate, Judge, Read: Role-Specialized Tool Agents\\
over a Knowledge-Point Index for Open-Domain Multimodal Document QA}

\author{Wenlin Lin \\
  Metask Lab \\
  \texttt{linwenlin.virhum@gmail.com}
  \And Fanzhe Wei \\
  Metask Lab \\
  \texttt{whyer1@gmail.com}}

\begin{document}
\maketitle

\begin{abstract}
We address question answering over large heterogeneous document collections: thousands of PDFs with tables and figures (open-domain M3DocVQA), chart-dense long reports (MMLongBench-Doc) and multi-document news corpora (MultiHop-RAG). A system must first \emph{find} the right passages and then \emph{read} them; current retrieval-augmented agents give one model both kinds of tools and let it schedule itself.
We present \sys{}, a pipeline separated by \emph{job}: knowledge construction into a hybrid-retrieval knowledge-point index; a \emph{Locator} agent that can only search; a code-scheduled \emph{bridge judge} that forces the missing second hop; a typed, size-bounded evidence pool that merges every search's ranking; a \emph{Deep Reader} agent that can only read text, images and tables; and a synthesizer with a format checker.
With a self-hosted Qwen3.6-35B-A3B (3B active), \sys{} reaches F1 47.5 on the full M3DocVQA (2,441 questions, 3,368 PDFs): +11.0 over the published M3DocRAG system (ColPali + Qwen2-VL-7B, 36.5) and +7.1 over one-shot retrieve-and-answer on the same index and model; 53.5 with DeepSeek-V4-Flash. On the multi-document MultiHop-RAG it reaches 73.5 accuracy, above the strongest published RAG and GraphRAG results (71.2 with Llama-3.1-70B) and the original GPT-4 baseline (56.0). On the single-document MMLongBench-Doc the pipeline lifts the same 3B-active model by 12.2 points, from 34.8 in one shot to 47.0, to the level of GPT-4o reading page images (46.3); only frontier vision-language models and RL-trained readers score higher.
Per-job ablations on the full benchmarks attribute the gain: the Deep Reader is worth 3.8, 9.8 and 0 points on the three corpora, tracking how visual each is; merging both agents into one with all tools and twice the rounds loses 1.9 F1 because it keeps searching instead of reading; a round-budget sweep shows a step at four Locator rounds and a plateau beyond, and query latency scales with model calls rather than corpus size.
To make these numbers checkable, we serve the three indexed benchmarks as read-only knowledge bases (free registration) and release all prompts and the runner code, so that retrieval recall can be reproduced without any model and answer accuracy with one's own: \url{https://github.com/metask-ai/knowforge-bench}.
\end{abstract}

\section{Introduction}

Answering a question over thousands of documents requires two different skills that pull in opposite directions.
Finding the right passages is an open-ended search problem: there is always another query to try.
Extracting the answer from a found passage is a closed reading problem: it requires opening a table, looking at a figure, or reading the paragraph around a hit, and it stops when the answer is in hand.
Agentic RAG systems give a single model both kinds of tools and a round budget and let the model schedule itself \cite{li2025searcho1,li2025webthinker,deepread2026}.
We find that when the two skills share one loop, the model systematically prefers the one with the immediate reward: it searches again rather than reading deeper.
In our controlled comparison, a single agent with every tool and eight rounds made 34\% more searches, read half as often, never opened a reading tool on 74\% of questions, and scored 1.9 F1 \emph{lower} than two agents with disjoint tools and four rounds each (\S\ref{sec:role}).

Prior systems take one of three routes. Single-pass retrieve-then-read systems such as M3DocRAG \cite{cho2024m3docrag} and HiKEY \cite{hikey2026} retrieve once and answer; they cannot recover the second document of a multi-hop question or open a figure they did not retrieve. Single-agent iterative systems (Search-o1 \cite{li2025searcho1}, WebThinker \cite{li2025webthinker}, DeepRead \cite{deepread2026}, SimpleDoc \cite{jain2025simpledoc}) let one model holding every tool decide when to search and when to read; we show below that such a model prefers to search. Multi-agent systems split roles by modality (MDocAgent \cite{han2025mdocagent}) or, most recently, by explore/refine/reflect stages on a single long document (MARDoc \cite{mardoc2026}, MAGE-RAG \cite{magerag2026}), but keep overlapping tool sets, leave the second hop to the model, and report no full-benchmark measurement of what each role is worth (Table~\ref{tab:compare}). Our step is to separate by \emph{job} with physically disjoint tools, force the second hop in code, hand over through a typed evidence pool, and quantify each part against the published systems on three full public benchmarks.

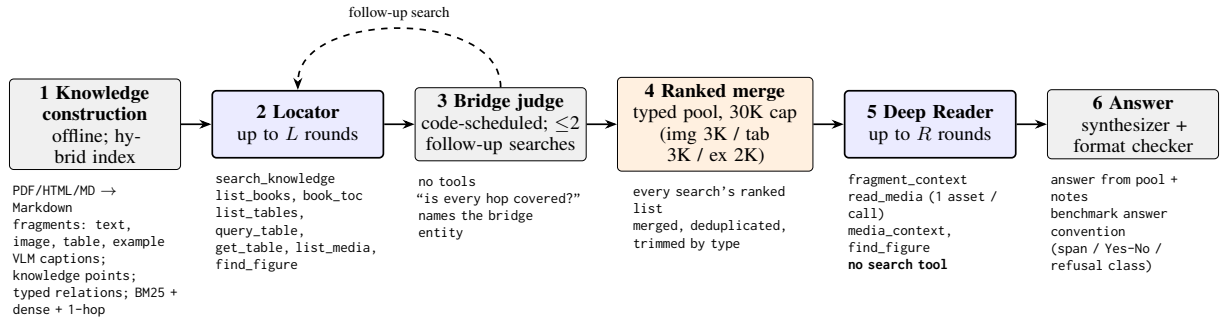
\begin{figure*}[t]
\centering
\resizebox{\textwidth}{!}{%
\begin{tikzpicture}[
  font=\small, node distance=4mm and 5mm,
  stage/.style={draw, rounded corners=2pt, align=center, minimum height=11mm, inner sep=3pt, text width=26mm},
  agent/.style={stage, fill=blue!8, thick},
  code/.style={stage, fill=gray!12},
  pool/.style={stage, fill=orange!12, text width=30mm},
  tools/.style={font=\scriptsize\ttfamily, align=left, text width=27mm, inner sep=1pt},
  arr/.style={-{Stealth[length=2mm]}, thick}
]
\node[code] (kc) {\textbf{1 Knowledge construction}\\ offline; hybrid index};
\node[agent, right=of kc] (loc) {\textbf{2 Locator}\\ up to $L$ rounds};
\node[code, right=of loc] (judge) {\textbf{3 Bridge judge}\\ code-scheduled; $\le$2 follow-up searches};
\node[pool, right=of judge] (pool) {\textbf{4 Ranked merge}\\ typed pool, 30K cap\\ (img 3K / tab 3K / ex 2K)};
\node[agent, right=of pool] (deep) {\textbf{5 Deep Reader}\\ up to $R$ rounds};
\node[code, right=of deep] (ans) {\textbf{6 Answer}\\ synthesizer + format checker};
\node[tools, below=2mm of kc] {PDF/HTML/MD $\to$ Markdown\\ fragments: text, image, table, example\\ VLM captions; knowledge points;\\ typed relations; BM25 + dense + 1-hop};
\node[tools, below=2mm of loc] {search\_knowledge\\ list\_books, book\_toc\\ list\_tables, query\_table,\\ get\_table, list\_media, find\_figure};
\node[tools, below=2mm of judge] {no tools\\ ``is every hop covered?''\\ names the bridge entity};
\node[tools, below=2mm of pool] {every search's ranked list\\ merged, deduplicated,\\ trimmed by type};
\node[tools, below=2mm of deep] {fragment\_context\\ read\_media (1 asset / call)\\ media\_context, find\_figure\\ \textbf{no search tool}};
\node[tools, below=2mm of ans] {answer from pool + notes\\ benchmark answer convention\\ (span / Yes-No / refusal class)};
\draw[arr] (kc) -- (loc);
\draw[arr] (loc) -- (judge);
\draw[arr] (judge) -- (pool);
\draw[arr] (pool) -- (deep);
\draw[arr] (deep) -- (ans);
\draw[arr, dashed] (judge.north) to[out=90,in=90] node[above, font=\scriptsize] {follow-up search} (loc.north);
\end{tikzpicture}}
\caption{The \sys{} pipeline. Stages 2 and 5 are agents with disjoint tool sets (blue); 1, 3 and 6 are code-scheduled (grey; 3 and 6 make one fixed model call each); 4 is the typed, size-bounded merge of all searches from stages 2--3 (orange).}
\label{fig:pipeline}
\end{figure*}

This paper argues that the unit of design for a document-QA agent should be the \emph{job}, not the round.
We describe \sys{}, a system whose pipeline is a chain of six jobs, each with its own tools and its own stopping rule (Figure~\ref{fig:pipeline}):
\begin{enumerate}
\item \textbf{Knowledge construction and hybrid retrieval} (offline). Documents become Markdown (OCR where needed), split into text, image, table and example fragments; images and tables get VLM descriptions; atomic \emph{knowledge points} are extracted and linked by typed relations; everything is indexed for BM25 and dense retrieval with bounded graph expansion.
\item \textbf{Locator.} An agent whose only tools search and enumerate (search, list books, table of contents, table lookup). It cannot read.
\item \textbf{Bridge judge.} A code-level step that asks, after locating, whether the evidence covers every hop of the question and, if not, issues one targeted follow-up search for the missing ``bridge'' document.
\item \textbf{Multi-list ranked merge.} All searches are merged into one 30K-token evidence pool that is \emph{typed}: text, images, tables and examples are ranked in separate sub-pools with their own caps, because text hits score higher than media hits and a single ranked list evicts figures and tables first.
\item \textbf{Deep Reader.} An agent whose only tools read (fragment context, read image, read table, media context). It has no search tool, so it cannot escape back into searching.
\item \textbf{Answer.} A synthesizer produces the answer from the pool and the reader's notes; a format checker normalises it to the benchmark's answer convention.
\end{enumerate}

Our contributions are:
\begin{itemize}
\item \textbf{A design principle and its realisation.} The unit of design for a document-QA agent is the job, not the round: Locator and Deep Reader are separate agents with physically disjoint tool sets, the second hop is forced by a code-level bridge judge rather than left to the model, and the hand-off between them is a typed, size-bounded evidence pool that merges every search's ranking so that figures and tables are not ranked out by text (\S\ref{sec:method}).
\item \textbf{Results against the published systems on three public benchmarks} with a self-hosted 35B mixture-of-experts model (3B active): full open-domain M3DocVQA F1 47.5 vs.\ 36.5 for M3DocRAG \cite{cho2024m3docrag} (+11.0) and 40.4 for one-shot retrieve-and-answer on the same index and model (+7.1); MultiHop-RAG accuracy 73.5 vs.\ 71.2 for the strongest published GraphRAG result (Llama-3.1-70B, \citealp{han2025ragvsgraphrag}) and 56.0 for the original GPT-4 baseline \cite{tang2024multihoprag}; on the single-document MMLongBench-Doc, +12.2 points on the same model (34.8$\rightarrow$47.0), bringing a 3B-active model level with GPT-4o reading page images (46.3, \citealp{ma2024mmlongbench}; \S\ref{sec:compare}). The same design on DeepSeek-V4-Flash reaches 53.5 on M3DocVQA (\S\ref{sec:main}, \S\ref{sec:model}, \S\ref{sec:transfer}).
\item \textbf{Per-job attribution of the gain}, all on the full benchmarks: the Deep Reader is worth 3.8 F1 on M3DocVQA, 9.8 points on MMLongBench-Doc and 0 on text-only MultiHop-RAG, tracking how visual the corpus is; role separation itself is worth 1.9 F1 against a single agent with every tool and twice the rounds; hybrid retrieval is worth 2.5 F1 inside the pipeline while typed graph expansion is neutral; a 12-configuration round-budget sweep shows a plateau below four Locator rounds, a step at four and nothing beyond (\S\ref{sec:role}--\S\ref{sec:index}).
\item \textbf{Cost and open verification.} Query latency is proportional to model calls and almost independent of corpus size (\S\ref{sec:efficiency}); the three indexed benchmarks are served as read-only knowledge bases (free account) with all prompts and runner code, so retrieval can be re-measured with no model and answers re-run with any model (\S\ref{sec:open}).
\end{itemize}

\section{Related Work}

\begin{table*}[t]
\centering\small
\resizebox{\textwidth}{!}{%
\begin{tabular}{@{}lccccccc@{}}
\toprule
System & Multi-round & Roles split by & Disjoint tool sets & Forced 2nd hop & Typed evidence pool & Open-domain full-set eval & Per-role ablation \\
\midrule
M3DocRAG \cite{cho2024m3docrag} & -- & -- & -- & -- & -- & \checkmark & -- \\
HiKEY \cite{hikey2026} & -- & -- & -- & -- & -- & \checkmark & $\circ$ \\
SimpleDoc \cite{jain2025simpledoc} & \checkmark & -- & -- & -- & -- & -- & $\circ$ \\
Search-o1 \cite{li2025searcho1} & \checkmark & stage (model-triggered) & $\circ$ & -- & -- & -- & $\circ$ \\
WebThinker \cite{li2025webthinker} & \checkmark & stage (model-triggered) & \checkmark & -- & $\circ$ & -- & $\circ$ \\
DeepRead \cite{deepread2026} & \checkmark & -- & -- & -- & -- & -- & $\circ$ \\
MDocAgent \cite{han2025mdocagent} & -- & modality & \checkmark & -- & -- & -- & \checkmark \\
MARDoc \cite{mardoc2026} & \checkmark & explore / refine / reflect & n/r & $\circ$ & $\circ$ & -- & $\circ$ \\
MAGE-RAG \cite{magerag2026} & \checkmark & -- & -- & -- & $\circ$ & -- & $\circ$ \\
\textbf{\sys{} (ours)} & \checkmark & \textbf{job} (code-scheduled) & \checkmark & \checkmark & \checkmark & \checkmark & \checkmark \\
\bottomrule
\end{tabular}}
\caption{Design comparison with the closest systems, from their published descriptions. $\circ$ = partial; n/r = not reported; ``open-domain full-set eval'' = results on the complete M3DocVQA corpus in one index (single-document systems marked --).}
\label{tab:compare}
\end{table*}

\paragraph{Multimodal document QA.}
M3DocVQA \cite{cho2024m3docrag} is the open-domain setting we target: 2,441 multi-hop questions over 3,368 Wikipedia-derived PDFs. The accompanying M3DocRAG system retrieves pages with ColPali \cite{faysse2025colpali} and answers with Qwen2-VL-7B in one pass (EM 31.4 / F1 36.5). HiKEY \cite{hikey2026} is the closest later open-domain system, with hierarchical parsing and hybrid retrieval but a single pass and no agent (EM 27.5 with a different reader). Scores of 60+ on M3DocVQA \cite{bookrag2025,hvmgraphrag2026} are on a 633-question subset and not comparable. MMLongBench-Doc \cite{ma2024mmlongbench} is the single-long-document complement (GPT-4o 46.3, RL-trained 8B agents 49.7 \cite{lin2026mmdocr1}, frontier vision-language models 57--62); MultiHop-RAG \cite{tang2024multihoprag} is a text-only multi-document benchmark whose strongest published RAG/GraphRAG numbers are 70--71 \cite{han2025ragvsgraphrag}. Visual-retrieval lines \cite{faysse2025colpali,yu2025visrag,tanaka2025vdocrag,ma2024dse} retrieve page images; we take the text route (OCR plus VLM descriptions of figures and tables), which \citet{dong2025mmdocir,dong2025mmdocrag} find more stable for open models.

\paragraph{Agentic and multi-stage RAG.}
Search-o1 \cite{li2025searcho1} and WebThinker \cite{li2025webthinker} separate a reasoning chain from a retrieval or exploration sub-process, but control passes by model tokens and the sub-process is not a reading agent; DeepRead \cite{deepread2026} gives one agent both a Retrieve and a ReadSection tool and is the natural foil for our physical tool isolation; MDocAgent \cite{han2025mdocagent} splits agents by modality; SimpleDoc \cite{jain2025simpledoc}, MAGE-RAG \cite{magerag2026}, MARDoc \cite{mardoc2026} and DocAgent \cite{docagent2025} iterate or split roles on a single long document. HiKEY, MARDoc and MAGE-RAG appeared as preprints in May--July 2026, shortly before our experiments, without released code; we compare designs in Table~\ref{tab:compare} and, for HiKEY, the published number. Our split is by job, tools are disjoint, the second hop is forced by code, and every job is ablated on open-domain corpora.

\paragraph{Graph-structured RAG.}
Our knowledge-point nodes descend from proposition retrieval \cite{chen2024densex,propraq2025}; zero-LLM similarity edges from HippoRAG \cite{gutierrez2024hipporag}, KGP \cite{wang2024kgp}, and LinearRAG \cite{linearrag2025}; bounded expansion from a retrieved seed from KG$^2$RAG \cite{zhu2025kg2rag}. Systematic evaluations \cite{han2025ragvsgraphrag,xiang2026graphragbench,gutierrez2025hipporag2} find that graph structure does not beat strong embedding RAG on factual QA; our index ablation agrees (\S\ref{sec:index}).

\section{Method}
\label{sec:method}

\subsection{Knowledge construction and hybrid retrieval}
Input files are converted to Markdown (marker for PDF with OCR fallback for scanned pages; a text extractor for HTML; native for Markdown). The Markdown is split into fragments typed as text, image, table, or example. Each image and table fragment receives a description from a vision-language model so that it participates in text retrieval; the original asset is kept for reading. From text fragments an LLM extracts atomic knowledge points (title, one-sentence summary, source fragment). Relations between knowledge points are typed (same-topic, extends, contrasts, supports, prerequisite) by recalling similar points with embeddings and letting an LLM assign the type only to recalled pairs. For M3DocVQA this yields 332K nodes and 299K edges (133K typed) over roughly 200K fragments.

Retrieval is hybrid: BM25 over an inverted index and dense retrieval over an HNSW index are fused, and the top-8 anchors are expanded by one hop along typed relations; each hit carries its adjacent-fragment context and any figures or tables adjacent to it. One retrieval round makes no LLM call.

\subsection{Locator}
The Locator receives the question, the corpus overview, and a tool set that can only \emph{find}: \texttt{search}, \texttt{list\_books}, \texttt{book\_toc}, \texttt{list\_tables}, \texttt{query\_table}, \texttt{get\_table}. It runs up to $L$ rounds and stops when it declares the evidence located. It never sees a reading tool.
\iflong
\paragraph{Why a separate agent rather than a phase.} An earlier version kept one agent and asked it, in the prompt, to spend its first rounds locating and the rest reading. The instruction had almost no effect: on M3DocVQA the agent spent its budget on searching and read figures on a small minority of image questions, and the single-agent baseline P0 in \S\ref{sec:role} reproduces the pattern with the current prompts (74\% of questions never see a reading tool). Searching is open-ended and rewarded immediately by new material; reading is bounded and costs a round without new hits. A model given both will search. The Locator therefore does not have the choice: its tool set contains nothing that reads.
\fi

\subsection{Bridge judge}
After the Locator stops, a fixed prompt asks the model whether the fragments found cover every entity the question chains through; if a hop is missing it names the missing entity and one follow-up query is issued for it. This is code-scheduled, bounded to one or two extra searches, and fires on roughly half of M3DocVQA questions. It exists because 78\% of the documents missed by the earlier version of the pipeline were second-hop ``bridge'' documents.
\iflong
\paragraph{Why code and not the model.} The Locator could in principle ask itself the same question. In practice it stopped as soon as it had material about the first entity: the name of the second document only becomes known after reading the first, and the agent that has just declared ``located'' does not reopen the search. Making the check a fixed step after the Locator's stop costs one short model call (about 300 output tokens) and removes the decision from the loop. On the full M3DocVQA the judge fires on 48\% of questions and issues 1.1 follow-up queries on average; on MultiHop-RAG, where every non-null question needs two to four articles, it is the single largest source of the 14.5-point rise in ``all evidence found''. A variant that let the judge also \emph{drop} evidence it thought irrelevant lost 3.1 F1 on the 300-question subset and was abandoned: a judge that only adds cannot hurt recall; one that also removes can.
\fi

\subsection{Multi-list ranked merge}
Every search made by the Locator and the bridge judge returns a ranked list. The lists are merged into one evidence pool that is handed, unchanged, to the Deep Reader and the synthesizer (the bridge judge sees its first 8K characters). Two design decisions matter.

\paragraph{Cap.} The earlier pipeline (M4) handed the reader only the first 12K tokens, which two searches already exceed: of the 1,063 fragments the M4 reader opened, 865 came from the first search, and only 24 of 1,710 multi-search questions ever read a later-search fragment. Replaying the M4 queries offline, a 12K cap keeps the answer-bearing fragment visible in 89.4\% of questions, 20K in 95.8\%, 30K in 99.0\%; we set 30K.

\paragraph{Typing.} When the pool is full, blocks must be evicted, and text fragments score systematically higher than image, table and example fragments (a caption matches a question less literally than a paragraph), so a single ranked list evicts media first. In the M4 run the answer-bearing fragment sat at rank 1 in 58.5\% of questions and within the top 8 in 91.7\%; the remaining 8.3\% sat at rank 9 or lower and were mostly figures, tables and examples (28\% of answer-bearing fragments were tables, 12\% images), exactly what table and figure questions need. The pool is therefore split into four sub-pools ranked separately (images 3K, tables 3K, examples 2K, text the remainder); within a sub-pool the top-3 hits of each search are protected and the lowest-scoring unprotected block is evicted first; expansion blocks and media attached to a text hit travel with it. In simulation the typed pool keeps the answer visible for 93.7\% of questions against 86\% for ``top-5 per search''.

\subsection{Deep Reader}
The Deep Reader receives the evidence pool (\S\ref{sec:pool}) and a tool set that can only \emph{read}: \texttt{fragment\_context}, \texttt{read\_media} (one image or table per call, answered by a VLM), \texttt{media\_context}, \texttt{find\_figure}, \texttt{list\_media}. It runs up to $R$ rounds. It has no search tool; an earlier version that kept search available saw the reader defect back to searching on 66 of 76 calls.
\iflong
\paragraph{What the reader receives.} The reader does not see the Locator's conversation. It receives the question and the typed evidence pool (\S\ref{sec:pool}) rendered as numbered material blocks, each search hit with its adjacent context and any figures or tables attached to it as ``[fig $N$]'' handles with page labels and fragment ids. Its tools take those ids: \texttt{fragment\_context} expands a hit to its neighbours or returns the full VLM description of a figure whose caption was truncated in the pool; \texttt{read\_media} sends one asset at a time with a single explicit question to the vision model and returns the answer verbatim (the prompt asks for the location of the evidence in the image and a one-line answer, and forbids using knowledge outside the image); \texttt{media\_context} returns the caption and body text on the page of a figure, because for an image fragment the ``adjacent'' fragments are other images, not its caption. Every result is appended to the same pool, so the synthesizer later sees both what was found and what was read.
\fi

\subsection{Answer}
A synthesizer writes the answer from the pool and the reader's notes. A format checker then rewrites it into the benchmark's convention (shortest exact span; ``Not answerable'' or ``Insufficient information'' where the benchmark has such a class; Yes/No). Benchmarks differ only in this last layer and in one rule paragraph in the agent prompts; the tools are identical.

\section{Experimental Setup}
\label{sec:setup}

\paragraph{Benchmarks.}
\textbf{M3DocVQA} \cite{cho2024m3docrag}: 2,441 questions, 3,368 PDFs, one shared index; EM and token F1; 980 multi-hop questions; evidence modality text/table/image. A stratified 300-question subset (\emph{mid300}) is used for sweeps.
\textbf{MMLongBench-Doc} \cite{ma2024mmlongbench}: 1,082 questions over 135 long PDFs (one index per document, as the task is closed-domain), 223 of them unanswerable; scored with the benchmark's model-judge protocol (we report the judge accuracy and, in parentheses, strict accuracy where the judge score equals 1).
\textbf{MultiHop-RAG} \cite{tang2024multihoprag}: 2,556 questions over 609 news articles, four types (inference, comparison, temporal, null); the official \texttt{has\_intersection} accuracy.

\paragraph{Models.}
Unless stated otherwise every text step runs on a self-hosted Qwen3.6-35B-A3B (FP8, 4$\times$RTX~4090, sglang); image reading uses the same model's vision input. DeepSeek-V4-Flash (API) is used for the model-axis experiment; Qwen2-VL-7B for the same-model reproduction of the M3DocRAG baseline.

\paragraph{Systems.}
\emph{S4}: one-shot search-and-answer on our index (one hybrid search, answer). \emph{M5}: the full pipeline with $L{=}4$, $R{=}4$. \emph{P0}: the \textbf{single-agent baseline} (ReAct-style): one agent holding every tool of both roles, 8 rounds, with the same judge, pool, synthesizer and checker as M5, so that the only difference is that one model schedules searching and reading itself. \emph{P1}: the \textbf{iterative-retrieval baseline}: Locator, judge, pool and synthesizer but no Deep Reader, i.e.\ multi-round search that never opens a figure, table or context window. \emph{$L{+}R$} variants change only the round caps. Index variants replace hybrid retrieval by dense-only (\emph{VEC}) or switch off graph expansion (\emph{S3}).

\paragraph{Runs and noise.}
Repeated full runs differ by at most 0.3 F1 with symmetric flips (179:184), so full-set differences above 1.4 F1 are outside noise; on mid300 the paired standard error is 2.0 F1. Every number in the paper comes from a single full run unless a repeat is stated. For paired comparisons we report the per-question win:loss count, a two-sided exact sign test on it, and a 95\% paired bootstrap interval of the mean difference (2,000 resamples).

\section{Results}
\label{sec:results}

\subsection{Main result and comparison with published systems}
\label{sec:main}\label{sec:compare}

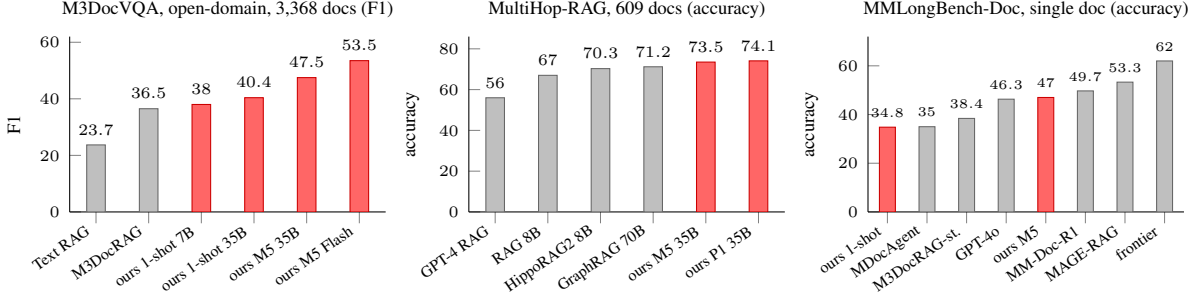
\begin{figure*}[t]
\centering
\begin{tikzpicture}
\begin{groupplot}[group style={group size=3 by 1, horizontal sep=1.1cm}, width=0.36\textwidth, height=3.9cm, ybar, /pgf/bar width=7pt, every axis plot/.append style={/pgf/bar shift=0pt},
  ymin=0, tick label style={font=\tiny}, label style={font=\scriptsize}, title style={font=\scriptsize, yshift=-3pt},
  x tick label style={rotate=35, anchor=east, font=\tiny}, nodes near coords, every node near coord/.append style={font=\tiny, /pgf/number format/fixed, /pgf/number format/precision=1},
  enlarge x limits=0.1, axis x line*=bottom, axis y line*=left]
\nextgroupplot[title={M3DocVQA, open-domain, 3{,}368 docs (F1)}, ymax=62, symbolic x coords={Text RAG,M3DocRAG,ours 1-shot 7B,ours 1-shot 35B,ours M5 35B,ours M5 Flash}, xtick={Text RAG,M3DocRAG,ours 1-shot 7B,ours 1-shot 35B,ours M5 35B,ours M5 Flash}, ylabel={F1}]
\addplot[fill=black!25, draw=black!60] coordinates {(Text RAG,23.7) (M3DocRAG,36.5)};
\addplot[fill=red!60, draw=red!80!black] coordinates {(ours 1-shot 7B,38.0) (ours 1-shot 35B,40.4) (ours M5 35B,47.5) (ours M5 Flash,53.5)};
\nextgroupplot[title={MultiHop-RAG, 609 docs (accuracy)}, ymax=86, symbolic x coords={GPT-4 RAG,RAG 8B,HippoRAG2 8B,GraphRAG 70B,ours M5 35B,ours P1 35B}, xtick={GPT-4 RAG,RAG 8B,HippoRAG2 8B,GraphRAG 70B,ours M5 35B,ours P1 35B}, ylabel={accuracy}]
\addplot[fill=black!25, draw=black!60] coordinates {(GPT-4 RAG,56.0) (RAG 8B,67.0) (HippoRAG2 8B,70.3) (GraphRAG 70B,71.2)};
\addplot[fill=red!60, draw=red!80!black] coordinates {(ours M5 35B,73.5) (ours P1 35B,74.1)};
\nextgroupplot[title={MMLongBench-Doc, single doc (accuracy)}, ymax=72, symbolic x coords={ours 1-shot,MDocAgent,M3DocRAG-st.,GPT-4o,ours M5,MM-Doc-R1,MAGE-RAG,frontier}, xtick={ours 1-shot,MDocAgent,M3DocRAG-st.,GPT-4o,ours M5,MM-Doc-R1,MAGE-RAG,frontier}, ylabel={accuracy}, /pgf/bar width=6pt, enlarge x limits=0.07, every node near coord/.append style={font=\fontsize{4.5}{5}\selectfont}]
\addplot[fill=red!60, draw=red!80!black] coordinates {(ours 1-shot,34.8)};
\addplot[fill=black!25, draw=black!60] coordinates {(MDocAgent,35.0) (M3DocRAG-st.,38.4) (GPT-4o,46.3)};
\addplot[fill=red!60, draw=red!80!black] coordinates {(ours M5,47.0)};
\addplot[fill=black!25, draw=black!60] coordinates {(MM-Doc-R1,49.7) (MAGE-RAG,53.3) (frontier,62.0)};
\end{groupplot}
\end{tikzpicture}
\caption{Published systems (grey) and ours (red; self-hosted 35B-A3B unless marked Flash). M3DocVQA and MultiHop-RAG are multi-document (full sets); MMLongBench-Doc is one document per question. Sources: Text RAG and M3DocRAG \cite{cho2024m3docrag}; HiKEY \cite{hikey2026} reports EM only (27.5 with a 7B reader vs.\ our 7B one-shot 31.7); GPT-4 RAG \cite{tang2024multihoprag}; RAG / HippoRAG2 (Llama-3.1-8B) and GraphRAG local (Llama-3.1-70B) \cite{han2025ragvsgraphrag}; MDocAgent, M3DocRAG-style (``M3DocRAG-st.'') and MM-Doc-R1 (RL-trained Qwen3-8B) \cite{lin2026mmdocr1}; GPT-4o reading pages \cite{ma2024mmlongbench}; MAGE-RAG \cite{magerag2026}; ``frontier'' = best 2026 public-leaderboard entry (Qwen3.6 Plus). On MMLongBench-Doc the pipeline lifts the same 35B-A3B model from 34.8 to 47.0 (+12.2), level with GPT-4o; the systems above it read charts with frontier or RL-trained vision models. Settings are not equalised; per-configuration numbers in Appendix~\ref{app:m3d}.}
\label{fig:compare}
\end{figure*}

Figure~\ref{fig:compare} places our results against the published systems; our claim is on the two multi-document benchmarks. On the full M3DocVQA the pipeline reaches EM 39.5 / F1 47.5 against 31.4 / 36.5 for M3DocRAG (44.6 / 53.5 with DeepSeek-V4-Flash); every later full-set system with a 7B reader reports EM $\le$ 27.5 \cite{hikey2026}, while our index with a 7B reader in one shot gives 31.7 / 38.0. On MultiHop-RAG, 73.5 is above the strongest published RAG and GraphRAG results (70.3 and 71.2). On the single-document MMLongBench-Doc the pipeline lifts the same 35B-A3B model by 12.2 points, from 34.8 in one shot to 47.0 (\S\ref{sec:transfer}), which puts a 3B-active open model level with GPT-4o reading page images (46.3); the systems above it are RL-trained readers or frontier vision-language models, whose advantage is in reading charts natively rather than in retrieval, which a single 50-page document does not require.

Three factors separate on M3DocVQA (Table~\ref{tab:m3d-all}). With index and one-shot procedure fixed, the 35B model gains 2.4 F1 over the 7B reader (238:181 flips). With the model fixed, the pipeline gains 7.1 F1 over one-shot (95\% CI $[5.5, 8.8]$; 447:255 at a 0.05 margin, sign test $p<10^{-12}$), raises the share of questions with all supporting documents found from 73.3\% to 84.6\% and with the answer span visible from 66.4\% to 77.2\%. Since our index with the 7B reader already exceeds the published visual-retrieval system, the gain over M3DocRAG is not a stronger reader.

\subsection{Which job earns the gain?}
\label{sec:role}

\begin{table}[t]
\centering\small
\resizebox{\columnwidth}{!}{%
\begin{tabular}{@{}lrrrrr@{}}
\toprule
 & F1 & Doc@all & Search/q & Read/q & No-read \\
\midrule
S4 one-shot & 40.4 & 73.3 & 1 & 0 & 100\% \\
P1: no Deep Reader & 43.7 & 83.3 & 3.52 & 0 & 100\% \\
P0: one agent, all tools, 8 rounds & 45.6 & 86.0 & 4.81 & 0.34 & 74\% \\
M5: two agents, 4+4 & \textbf{47.5} & 84.6 & 3.58 & 0.71 & 58\% \\
\bottomrule
\end{tabular}}
\caption{Role ablation on full M3DocVQA. Search/q and Read/q are mean tool calls per question; No-read is the share of questions on which no reading tool was ever called. P0 finds documents slightly better than M5 and answers worse.}
\label{tab:role}
\end{table}

Table~\ref{tab:role} is a ladder of reading effort. P1 (iterative retrieval, nobody reads) is +3.3 over one-shot, almost all retrieval. P0 (single agent, every tool, twice the rounds) searches 34\% more than M5, reads half as often, never opens a reading tool on 74\% of questions, and lands 1.9 F1 below M5 (215:157; CI $[0.5, 3.4]$, $p=0.005$) despite finding slightly more documents (86.0 vs 84.6). M5 always runs a reader and gains on every modality bucket (image 29.5 vs 27.8, table 40.0 vs 37.8, text 60.5 vs 58.9). Reads per question go 0$\rightarrow$0.34$\rightarrow$0.71 across P1$\rightarrow$P0$\rightarrow$M5 and F1 follows monotonically. Removing the Deep Reader from M5 costs 3.8 F1 (149:247; CI $[2.3, 5.3]$, $p<10^{-5}$), concentrated on table questions ($-5.5$) and on questions whose supporting document was already the top-1 hit ($-6.5$): the document was found, nobody read it.

\subsection{The evidence pool: cap and typing}
\label{sec:pool}
M5 differs from M4 (Table~\ref{tab:m3d-all}) in the bridge judge and the pool: +1.9 F1 on the full set (206:155; CI $[0.5, 3.3]$, $p=0.003$), +3.6 on documents found, with table +2.3, image +1.7 and multi-hop +2.2. The cap itself is neutral: the 811 questions whose materials exceed 30K gain as much as the 1,630 that do not (28.9$\rightarrow$30.6 vs 53.9$\rightarrow$56.0); the pool's value is that the reader sees everything up to 30K instead of the first 12K. Typing is what makes the larger pool safe: on the 300-question subset a single ranked list with cosine eviction (pool v1) scored 23.3 F1 on the questions that hit the cap, against 30.6 for the typed pool, because tables were evicted below text of the same search.

\subsection{Round budget: a step, then a plateau}
\begin{figure}[t]
\centering
\begin{tikzpicture}
\begin{axis}[width=0.94\columnwidth, height=4.6cm, xtick={0,...,6}, xticklabels={0+0,1+1,2+2,3+3,4+4,5+5,6+6},
  xlabel={round caps: Locator $+$ Reader}, ylabel={answer F1 (mid300)}, ymin=40, ymax=54, axis y line*=left,
  tick label style={font=\scriptsize}, label style={font=\scriptsize}, legend style={font=\tiny, at={(0.02,0.98)}, anchor=north west, draw=none, fill=none},
  grid=major, grid style={opacity=0.25}]
\addplot[red!70!black, thick, mark=*, mark size=1.6pt] coordinates {(0,43.1)(1,46.4)(2,46.7)(3,46.2)(4,50.9)(5,49.2)(6,50.6)};
\addlegendentry{answer F1}
\addplot[red!70!black, only marks, mark=x, mark size=2.4pt] coordinates {(4,47.6)(1,48.8)(3,48.6)(4,50.3)};
\addlegendentry{single-role cuts}
\node[font=\tiny, red!70!black, anchor=west] at (axis cs:4.08,47.6) {4+1};
\node[font=\tiny, red!70!black, anchor=west] at (axis cs:1.08,48.8) {1+4};
\node[font=\tiny, red!70!black, anchor=west] at (axis cs:3.08,48.6) {3+4};
\node[font=\tiny, red!70!black, anchor=south west] at (axis cs:4.08,50.3) {4+3};
\end{axis}
\begin{axis}[width=0.94\columnwidth, height=4.6cm, xtick=\empty, ymin=68, ymax=94, axis y line*=right, axis x line=none,
  ylabel={retrieval (\%)}, tick label style={font=\scriptsize}, label style={font=\scriptsize},
  legend style={font=\tiny, at={(0.98,0.02)}, anchor=south east, draw=none, fill=none}]
\addplot[black!70, thick, dashed, mark=square*, mark size=1.4pt] coordinates {(0,71.0)(1,77.3)(2,80.7)(3,82.0)(4,83.8)(5,83.7)(6,84.0)};
\addlegendentry{documents found}
\addplot[black!45, thick, dotted, mark=triangle*, mark size=1.6pt] coordinates {(0,83.0)(1,87.6)(2,90.1)(3,90.5)(4,91.5)(5,91.1)(6,91.7)};
\addlegendentry{union recall R@all}
\end{axis}
\end{tikzpicture}
\caption{Round-cap sweep on mid300 (Table~\ref{tab:budget}). Retrieval (right axis) rises and saturates; answer F1 (left axis) is flat for caps 1--3, steps up at 4+4 and does not rise at 5+5 or 6+6. Single-role cuts show the Reader cap matters below 3 and the Locator cap below 4.}
\label{fig:budget}
\end{figure}
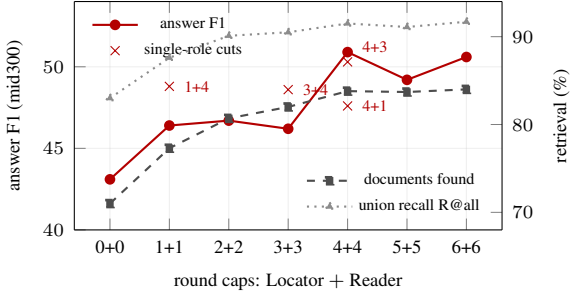
\label{sec:budget}

\begin{table}[t]
\centering\small
\resizebox{\columnwidth}{!}{%
\begin{tabular}{@{}lrrrrr@{}}
\toprule
$L{+}R$ & F1 & Doc found & R@all & Loc calls & Read calls \\
\midrule
0+0 (S4) & 43.1 & 71.0 & 83.0 & -- & -- \\
1+0 & 44.7 & 78.0 & 87.7 & 2.11 & 0 \\
1+1 & 46.4 & 77.3 & 87.6 & 2.09 & 0.24 \\
2+2 & 46.7 & 80.7 & 90.1 & 2.45 & 0.38 \\
3+3 & 46.2 & 82.0 & 90.5 & 3.08 & 0.48 \\
\textbf{4+4} (3 runs) & \textbf{50.9} & 83.8 & 91.5 & 3.45--3.58 & 0.67--0.73 \\
5+5 & 49.2 & 83.7 & 91.1 & 3.98 & 0.85 \\
6+6 & 50.6 & 84.0 & 91.7 & 4.45 & 0.90 \\
\bottomrule
\end{tabular}}
\caption{Round-cap sweep on mid300 (Qwen3.6). Caps are upper bounds; the last two columns are mean calls actually made. Single-role cuts (4+1, 1+4, 3+4, 4+3) are in Figure~\ref{fig:budget} and Appendix~\ref{app:usage}.}
\label{tab:budget}
\end{table}

Table~\ref{tab:budget} and Figure~\ref{fig:budget} show two different curves. Retrieval improves smoothly and saturates: documents found 71$\rightarrow$77$\rightarrow$81$\rightarrow$82$\rightarrow$84, then 84 at any larger cap. Answer quality is flat at 46.2--46.7 for caps 1+1 through 3+3, steps to 50.9 at 4+4 (three repeats, range 0.4; the 4.4-point gap to the 1--3 plateau is well outside the $\pm$2 paired error), and stays at 49.2--50.6 for 5+5 and 6+6 while calls and latency keep growing (+25\% time from 4+4 to 6+6). Single-role cuts locate the step: Locator 4$\rightarrow$3 or 4$\rightarrow$1 costs about 2 F1; Reader 4$\rightarrow$1 costs 3.3, 4$\rightarrow$3 nothing. With at most three hops, three effective searches plus one re-search is the smallest budget that covers the worst case.\iflong{} The model uses only part of any cap (mean Locator calls 3.5 at a cap of 4; the Reader is idle on 55--76\% of questions).\fi

\subsection{Index factors inside the pipeline}
\label{sec:index}
Keeping M5 fixed and changing only the index: dense-only retrieval drops F1 to 45.0 ($-2.5$, 155:226; CI $[-4.0, -1.1]$, $p=0.007$) with first-search R@8 78.8$\rightarrow$72.5; switching off typed graph expansion gives 47.0 ($-0.5$, 180:201; $p=0.56$), first-search recall slightly up and answer visibility slightly down. Hybrid retrieval is a real component of the pipeline; graph expansion is neutral here, in line with recent evaluations of graph RAG on factual QA \cite{han2025ragvsgraphrag,xiang2026graphragbench}, and we keep it only because it costs 45\,ms per round.

\subsection{Model axis}
\label{sec:model}
On the same 300 questions, one-shot$\rightarrow$M5 gains 7.8 F1 with Qwen3.6 (43.1$\rightarrow$50.9) and 6.1--6.9 with DeepSeek-V4-Flash (47.2$\rightarrow$54.1 / 53.3). The document-found rate is identical for both models in one shot (71.0) and rises to 82--84 for both in the pipeline: the retrieval layer is model-independent, and the pipeline's gain is a shift of about the same size on both. On the full set Flash reaches 53.5 (Figure~\ref{fig:compare}), 6.0 above Qwen3.6 at the same pipeline, so the design does not need a weak model to show its effect. An earlier version of the pipeline run in full with DeepSeek-V4-Pro reached F1 51.1 against 50.8 for Flash on the same version, so the two API models sit in one band and we kept the cheaper one. At the small end, Qwen2-VL-7B driven through the same multi-round protocol did worse than its own one-shot (mid300 37.4 vs 40.6) and Qwen3-VL-8B only tied it (41.5--42.3 vs 41.3): the pipeline found them 9 more points of documents, but they could not convert the extra evidence. The protocol assumes a model that can follow a tool-call format across rounds; below that threshold the gain is retrieval-only.

\subsection{Transfer to other corpora}
\label{sec:transfer}

\begin{table*}[t]
\centering\small
\begin{tabular}{@{}lrrrr@{}}
\toprule
 & S4 & 1+1 & M5 4+4 & P1 no reader \\
\midrule
\multicolumn{5}{@{}l}{\emph{MMLongBench-Doc} (1,082 q; judge acc., strict in parentheses)} \\
All & 34.8 (32.8) & 41.5 (38.8) & \textbf{47.0} (44.1) & 37.2 (35.0) \\
Answerable (859) & 16.6 & 24.9 & \textbf{31.7} & 19.9 \\
Unanswerable (223) & 95.1 & 92.4 & 91.9 & 93.3 \\
Table (143) & 22.4 & 37.8 & \textbf{46.9} & 26.6 \\
Chart (107) & 15.0 & 21.5 & \textbf{32.7} & 16.8 \\
Figure (185) & 16.8 & 23.8 & \textbf{29.2} & 19.5 \\
Pure text (131) & 25.2 & 28.2 & 32.8 & 30.5 \\
\midrule
\multicolumn{5}{@{}l}{\emph{MultiHop-RAG} (2,556 q; official accuracy)} \\
All & 68.2 & 70.8 & 73.5 & \textbf{74.1} \\
Inference (816) & 94.0 & 93.9 & 95.2 & 96.1 \\
Comparison (856) & 44.3 & 50.1 & 53.7 & 54.4 \\
Temporal (583) & 51.5 & 54.9 & 59.5 & 60.2 \\
Null (301) & 98.3 & 97.7 & 98.0 & 97.0 \\
All evidence found & 50.6 & 60.0 & 65.1 & 65.6 \\
False refusals & 18.1\% & 14.1\% & 9.8\% & 10.6\% \\
\bottomrule
\end{tabular}
\caption{Transfer. Same tools, same prompts up to one benchmark-specific rule paragraph and the answer-format checker. On the chart-heavy MMLongBench-Doc the Deep Reader is the largest single factor; on the text-only MultiHop-RAG it contributes nothing and the cheaper P1 ties M5.}
\label{tab:transfer}
\end{table*}

Table~\ref{tab:transfer} runs the same four configurations on two corpora of different structure. On MMLongBench-Doc, S4 has no image tool and is a reference row; the fair comparison is across the three same-tool configurations. Going from 1+1 to 4+4 gains 5.5 points (CI $[3.4, 7.5]$, $p<10^{-5}$); removing the Deep Reader loses 9.8 (136:38; CI $[7.6, 12.2]$, $p<10^{-14}$) and P1 even falls below 1+1 (55:96): on a corpus of charts and tables, searching more without reading does not help. The 12.2-point gain brings the 3B-active model level with GPT-4o reading page images (46.3 accuracy / 44.9 F1 against our 47.0 / 41.4); frontier vision-language models that read the whole document natively score 57--62 (\S\ref{sec:compare}), an advantage in chart reading rather than retrieval, which a single 50-page document does not need. On MultiHop-RAG the ordering changes: P1 ties M5 (74.1 vs 73.5, 115:129; CI of the difference $[-1.8, 0.6]$, $p=0.41$), while M5 beats one-shot by 5.4 (243:106, $p<10^{-12}$) and 1+1 by 2.7 (174:104, $p<10^{-4}$). Every article is short and lands whole in the 30K pool, so there is nothing for a reader to open; the gain comes from the Locator's extra rounds and the bridge judge, which raise the share of questions with all evidence found from 50.6\% to 65\% and lift the evidence-hungry comparison and temporal types by 9--10 points. The strongest published numbers are 70.3 (HippoRAG2, Llama-3.1-8B) and 71.2 (GraphRAG local search, Llama-3.1-70B) \cite{han2025ragvsgraphrag}, the original GPT-4 baseline 56.0 \cite{tang2024multihoprag}; comparisons are indicative only, since retrieval units and contexts differ.

Across the three corpora the Deep Reader is worth +3.8, +9.8 and 0, tracking how much evidence lives in figures and tables, while Locator plus judge is positive everywhere: jobs can be switched on per corpus.
\iflong
\paragraph{MMLongBench-Doc in detail.} The answerable questions split 499 single-page and 360 cross-page. Every configuration is far better on single-page questions (M5 43\% vs 25\% judge accuracy, Appendix~\ref{app:transfer}); on cross-page questions the best configuration reaches 25\% and the one-shot baseline 11\%, so the pipeline helps there too, but merging evidence that lives on two pages, typically a chart on one and its legend or a total on another, remains the main weakness. Refusals dominate the error mass: of the 859 answerable questions, one-shot refuses 475 and M5 274, and the 201-question difference is a larger part of M5's gain than the questions it turns from wrong to right. Reading is the mechanism: M5 calls \texttt{read\_media} on 0.50 questions on average and never on 636 of 1,082, against 0.27 and 788 for 1+1 and none for P1; the Chart bucket, where the answer is a value read off a plot, moves from 15 to 33. Earlier versions of the pipeline on the same questions scored 46.2, 44.3 and 45.4 with Qwen3.6 and 49.0 and 48.7 with DeepSeek-V4-Flash, so the current version sits at the top of the band rather than being an outlier; GPT-4o reading page images sits at 46.3 accuracy / 44.9 F1, MM-Doc-R1 at 49.7 / 46.1 with an RL-trained Qwen3-8B \cite{lin2026mmdocr1}, and frontier vision-language models at 57--62 on 2026 leaderboards.
\paragraph{MultiHop-RAG in detail.} The four question types behave very differently. Inference questions are at 94--96\% under every configuration even though they have the lowest all-evidence recall (R@all 73.0, Appendix~\ref{app:transfer}): the answer is reachable without every evidence article. Null questions, which have no answer in the corpus, are at 97--98\%: the conservative answer convention works. The gain is entirely in comparison and temporal questions, which need two or three articles and a judgement across them: 44$\rightarrow$54 and 52$\rightarrow$60 from one-shot to the pipeline. The mechanism is visible in retrieval: first-search R@8 actually falls from 73.9 to 61.0 when the Locator is allowed several rounds, because its first query becomes a targeted sub-question rather than the whole question, but the union over all searches rises from 76.7 to 84.2 and the share of questions with every evidence article found from 50.6 to 65.1. False refusals fall from 409 (18.1\% of non-null questions) to 220 (9.8\%); 149 questions that one-shot refused are answered correctly by the pipeline. The original GPT-4 RAG baseline is 56.0 (89.0 with gold evidence) and the strongest published numbers are 70.3 for HippoRAG2 with Llama-3.1-8B and 71.2 for GraphRAG local search with Llama-3.1-70B \cite{han2025ragvsgraphrag}; a third of the remaining gap to the gold ceiling is reasoning over the evidence rather than finding it. Two prompt variants that pushed the model to always conclude Yes or No when two articles agree reduced refusals further but lost 12.9 points on a 31-question smoke set by forcing conclusions on questions whose gold answer was a name; we kept the conservative prompt.
\fi

\subsection{Efficiency}
\label{sec:efficiency}
\paragraph{Query latency.} Measured serially, one-shot takes 4.6\,s (p50) and the full pipeline 27.3\,s with self-hosted Qwen3.6 (6.0 and 31.9\,s with the Flash API\iflong; Table~\ref{tab:latency}\fi). The pipeline makes about nine model calls of 3--4.6\,s each; its four retrieval rounds, 0.12\,s each end to end, are 2\% of the total\iflong{} (timed one stage at a time: embedding 40\,ms, BM25 18, HNSW 8, graph expansion 45, context 30)\fi.
\iflong
\begin{table}[t]
\centering\small
\resizebox{\columnwidth}{!}{%
\begin{tabular}{@{}lrrrrr@{}}
\toprule
System (serial, 100 q) & p50 / p90 & first event & answer start & calls & F1 \\
\midrule
Qwen3.6 one-shot & 4.6 / 19.6\,s & -- & 1.7\,s & 2 & 31.1 \\
Qwen3.6 pipeline M5 & 27.3 / 65.3\,s & 1.2\,s & 23.5\,s & $\approx$9 & 43.2 \\
Flash API one-shot & 6.0 / 13.4\,s & -- & 2.4\,s & 2 & 39.3 \\
Flash API pipeline M5 & 31.9 / 47.2\,s & 1.7\,s & 27.6\,s & $\approx$9 & 50.5 \\
\bottomrule
\end{tabular}}
\caption{Serial latency on the first 100 mid300 questions, one question at a time, no other load. ``First event'' is the first tool call (what a user would see first); ``answer start'' is the first synthesizer token. The 27\,s of the pipeline is nine model calls of 3--4.6\,s; the API model is slower per call but has a tighter p90 because the self-hosted server is shared with production traffic.}
\label{tab:latency}
\end{table}
\fi
\paragraph{Corpus scale.} Growing the corpus 42$\times$ (80$\rightarrow$3,368 PDFs; 3.7K$\rightarrow$200K fragments) raises one retrieval round by 7\% (0.113$\rightarrow$0.121\,s). The cost of scale is accuracy, not time: on the same 54 questions one-shot R@8 falls 99.4$\rightarrow$78.1 and F1 38.1$\rightarrow$31.9 as distractors grow, which is why the second-hop search exists.
\paragraph{Index construction.} It is a one-time cost: 138M model tokens, 202K embeddings and 17K OCR pages for M3DocVQA (about 12 hours), 20M tokens for MMLongBench-Doc, 3.1M for MultiHop-RAG (Appendix~\ref{app:cost}).

\section{Analysis}
\label{sec:analysis}

\paragraph{Where the remaining errors are.}
Of the M3DocVQA questions whose supporting document is the first hit yet F1 $<0.5$, a 100-question sample classified by DeepSeek-V4-Flash and independently re-checked by a second LLM reviewer with the full traces (88\% agreement; Appendix~\ref{app:errors}) attributes 31 to misreading or not opening an image, 23 to answer granularity the token-F1 metric penalises (``rifle'' vs ``gun''), 19 to the wrong field in the right document, 13 to refusing with the answer present in the text, 9 to the wrong table row or column and 5 to multi-hop assembly. With retrieval at 92\% recall, more than half of the residual loss is not retrieval.

\iflong
\paragraph{A worked example.} Appendix~\ref{app:example} traces one two-document question through all four configurations. \emph{``Which driver with a time/retired of +1 lap in the Trenton Twin Indy of 1979 has won the Indy 500 the most times?''} (gold: Rick Mears). The race page's prose names the winners; the ``+1 lap'' finishers exist only in the results table, stored as an image. One-shot answers from the prose (``Al Unser'', wrong, 2\,s). The iterative-retrieval baseline P1 searches the race page four times, and its bridge judge, reasoning from the prose, asks for the Indy~500 wins of the wrong three drivers; the follow-up search does retrieve the \emph{List of Indianapolis 500 winners}, but with nobody reading the table the answer is again ``Al Unser''. The full pipeline runs the same Locator and judge, then the Deep Reader opens the results table with \texttt{read\_media} and reads the four +1-lap finishers (Dallenbach, Mears, Sneva, Ongais); with the winners list already in the pool the synthesizer answers ``Rick Mears'' (46\,s under the 4-way harness). The single-agent P0 also solves this question, by reading the table right after its first search; the point of Table~\ref{tab:role} is that on average it does so far less often.
\fi
\paragraph{False refusals.}
On the two benchmarks with an unanswerable class the pipeline still refuses too often on answerable questions (MultiHop-RAG 9.8\%, MMLongBench-Doc 274 of 859), although the multi-round configuration halves the rate relative to one-shot (18.1\%$\rightarrow$9.8\%; 475$\rightarrow$274). The residual is the price of the conservative answer convention; relaxing it trades the unanswerable bucket one for one.

\paragraph{Honest boundaries.}
(i) Cross-page questions on MMLongBench-Doc stay low under every configuration (10--22\% of 358); merging evidence across pages is a shared weakness. (ii) Graph expansion is neutral here.\iflong{} (iii) The round-cap plateau was measured on one model whose loop shows the remaining budget; the reader stops early when little remains rather than being cut off.\fi{} (iv) Closed-book scores on M3DocVQA are non-trivial for strong models (Flash 43.0, Qwen3.6 32.2) because the corpus is Wikipedia; we therefore report gains over the same model's one-shot rather than over closed-book.

\iflong
\section{Discussion}
\label{sec:discussion}
\paragraph{Which model can drive this pipeline.} The protocol asks the model to emit one JSON object per round, to stop with a placeholder rather than an answer, and to keep a thought to one or two sentences. Qwen3.6-35B-A3B and DeepSeek-V4-Flash follow it; Qwen2-VL-7B and Qwen3-VL-8B do not reliably: on the 300-question subset 113 of 300 runs of the 7B ended in a forced stop of both agents, and their multi-round scores (37.4 and 41.5--42.3) did not beat their own one-shot (40.6 and 41.3) even though the pipeline found them 9 more points of documents. The extra evidence was there and could not be converted. Prompt choice is also model-dependent: the official M3DocRAG answer template was the best of three prompts for the 7B model and the worst for the 35B model ($-3.3$ F1 on the subset, $-1.4$ on the full set inside the pipeline), and a ``positive phrasing'' rewrite that removed every prohibition from the prompts lost 4.6 on the subset. We therefore did not tune prompts per model, and report the protocol threshold as a limitation: below it, the gain of the design is retrieval-only.
\paragraph{Round budgets are upper bounds.} The sweep in \S\ref{sec:budget} is often read as ``more rounds help until four''. The usage data (Appendix~\ref{app:usage}) says something narrower: from a cap of 3 upward the Locator reaches its cap on only 44--57\% of questions, the Reader makes no call on 55--76\%, and the amount of reading depends only on the Reader's own cap (0.24, 0.38, 0.48, 0.67--0.73, 0.85, 0.90 calls per question for caps 1--6), not on the Locator's. Neither cap alone explains the step from 3+3 to 4+4 (+4.7 F1): with the other cap at 4, the Locator's fourth round is worth 2.3 and the Reader's 0.6, although the Reader's fourth round raises reads by half (0.47$\rightarrow$0.67--0.73) while its cap-hit rate moves only from 23\% to 17--18\%. Raising both caps further raised reads again and F1 not at all. The model appears to stop early when it can see that little budget remains, and to read more when it has room, up to a point. This is an observation about one model in a loop that displays the remaining budget, not a general law; the threshold will move with the model and the loop design.
\paragraph{What the graph layer is for.} The knowledge-point graph (typed relations between atomic statements) was built for browsing and summarisation in the product, and we tested whether it also helps retrieval. Inside the pipeline, expanding the top hits along typed edges is neutral (+0.5 F1, $p=0.56$): it slightly raises answer visibility on single-hop questions and slightly lowers first-search recall on the rest by displacing BM25 hits from a fixed-size list. Hybrid lexical-plus-dense retrieval, by contrast, is worth 2.5 F1 inside the pipeline while only 0.5 in one shot: multi-round search converts better ranking into found documents, because the Locator decides what to read from the first few hits. We report the neutral result rather than dropping the component, since its cost is 45\,ms per round and its value may differ on corpora with denser cross-document relations than a sample of Wikipedia.
\paragraph{Deployment.} The same six-job chain is the question-answering path of the product the corpora were indexed with; the benchmark harness differs from the product only in the answer-format checker and one rule paragraph per benchmark. The design decision that carried over most directly is the disjoint tool set: in production the reader agent has no search tool for the same reason as in the benchmark. The 27-second latency is acceptable for a research assistant and not for a chat interface; the first visible event at 1.2\,s and the streamed synthesis are what make it usable.
\paragraph{Outlook.} The pipeline's gain is largest where the model is weakest at scheduling itself, and the jobs are simple enough that each could be served by a smaller or specialised model: a Locator needs query reformulation and stopping, a Reader needs tool discipline and vision, a judge needs one structured decision. Whether a set of small, job-specialised agents can match a large general model on this task is the natural next experiment; the per-stage model overrides in the released runner are there to make it easy to run.

\fi
\section{Open Verification}
\label{sec:open}
The three benchmark indexes are served read-only by the product's retrieval API to anyone with a free account (an API key plus the index identifiers). We release the identifiers, fetch scripts for the official question sets, the scoring scripts, all prompts, and a reference runner that takes one configuration file (model endpoint, API address, index identifier) and produces a scored run. A third party can recompute document recall from the search endpoint alone, run the reference agent with their own model, or score their own agent against the same indexes.\iflong{} The served indexes were rebuilt from the same converted documents and extracted knowledge points as our runs, with figure descriptions and typed relations regenerated by the same model family; on the 299-question MultiHop-RAG subset with the same Qwen3.6 endpoint, the released runner scores 73.9 against 71.9 for our in-process harness (16:10 flips), inside the run-to-run band.\else{} The served indexes are rebuilt from the same documents; the runner matches our harness within run-to-run noise.\fi{} Our model endpoint is not exposed and we redistribute neither documents nor raw outputs; images are read by downloading the asset into the third party's own model. Repository: \url{https://github.com/metask-ai/knowforge-bench}.

\section{Conclusion}
Splitting a document-QA agent by job, with disjoint tools and code-scheduled hand-offs, beats one-shot retrieval by 7 F1 on the full M3DocVQA and the published systems on both multi-document benchmarks, and makes each job's value measurable: the reader matters where evidence is visual, the locator and forced second hop everywhere, extra rounds and the graph layer nowhere.

\section*{Limitations}
The main line runs on one self-hosted model, but the pipeline is validated in full on two model families: Qwen3.6-35B-A3B and DeepSeek-V4-Flash (F1 47.5 and 53.5 on M3DocVQA; 47.0 and 49.0 on MMLongBench-Doc), with DeepSeek-V4-Pro run in full on the earlier pipeline version (51.1, in the same band as Flash). Smaller open models (Qwen2-VL-7B, Qwen3-VL-8B) were run on the 300-question subset only, where the multi-round protocol did not beat their own one-shot; the claims therefore hold for models that can follow a multi-round tool protocol, and the round-cap plateau in particular was measured only on Qwen3.6. M3DocVQA is Wikipedia-derived and partially memorised by large models, which is why we anchor on one-shot rather than closed-book. MMLongBench-Doc is scored by a model judge (we use our own Qwen3.6 as judge, following the benchmark's protocol) and judge choice moves scores by 2--3 points. Latency figures are for one hardware configuration and one serial client. We did not tune prompts per benchmark beyond one rule paragraph and the format checker, and we report a prompt variant that helped on a smoke test and hurt on the full set as a negative result in the appendix.

\section*{Ethics Statement}
The benchmarks are used under their licences (MMLongBench-Doc data CC BY-NC 4.0, research use; MultiHop-RAG ODC-BY; M3DocVQA from Wikipedia content). The public indexes are labelled for research use, require a free account and are rate-limited. No private documents were used.

\bibliography{references}

\appendix
% ---------- Appendix (input from main.tex after \appendix) ----------

\section{Agent prompts}
\label{app:prompts}

The agent-side prompts are written in Chinese because the deployed system serves Chinese users; the answer-side prompts (guardrail appended to the question, bridge judge, format checker) are in English because the benchmarks are English. We give faithful English renderings here; the verbatim originals are released with the code. Every configuration in the paper uses these texts unchanged except for the one benchmark-specific rule paragraph and the format checker noted below.

\paragraph{Shared loop protocol (both agents).}
The model answers each round with a JSON object \texttt{\{"thought": ..., "tool": ..., "args": ...\}} or stops with \texttt{\{"thought": ..., "final": "ok"\}}. The rules state that the thought is one or two sentences, that the agent must never write the answer text in the loop (it is generated by the synthesizer from the collected materials and any answer written here is discarded), and that the agent should stop as soon as the materials suffice. Every tool result is followed by a one-line reminder of the same stopping rule; we found in-context reminders far more effective than rules at the top of the system prompt for the models used.

\paragraph{Locator rules (English rendering).}
\begin{quote}\small
You are in the \textbf{locating stage}: your only task is to find the books and fragments that hold the answer; do not dig into details. Use \texttt{search\_knowledge} to search; use \texttt{list\_books} to see which books the library has and \texttt{book\_toc} to see a book's structure. As soon as a search result clearly carries the answer (or results largely repeat what you already have), stop with \texttt{\{"thought": ..., "final": "ok"\}}; details are handled in the next stage. Do not try to answer the question in this stage.

For \textbf{enumeration or counting questions} (``how many figures / tables / photos does this document have'') do not count by searching: use \texttt{list\_media} to get the full list of figures and tables (filterable by kind or page); the list is the material. For questions that \textbf{name a figure or table number} (``Figure 5'', ``Table 3'') use \texttt{find\_figure} to locate that page and its media fragment rather than guessing with search. When the question has \textbf{two or more parts or compares two sides}, both sides must be located before you stop; one side alone does not count as located.
\end{quote}

\paragraph{Bridge judge (verbatim, English).}
\begin{quote}\small\ttfamily
You audit whether collected materials are sufficient to answer a question that may need TWO hops. A ``bridge entity'' is a person/place/work/organization that the materials reveal as the answer to the FIRST part of the question, while the question then asks something ABOUT that entity (its county, founder, release year, genre, ...). That entity's own document must be retrieved to answer. Decide: If the materials already contain the final answer, or the question is single-hop, output need\_more=false. If a bridge entity is identified but its own document has NOT been retrieved yet (the materials only mention it in passing), output need\_more=true with 1-2 short search queries that retrieve that entity's document plus the asked attribute (e.g. ``Bay Lake Florida county''). Also output need\_more=true if the question names a second document/title that none of the searched queries targeted. Never repeat a query already searched. Output JSON only: \{"need\_more": true|false, "bridge\_entities": [...], "queries": [...], "reason": "<one sentence>"\}
\end{quote}
The judge receives the question, the list of queries already searched, and the first 8K characters of the collected materials. Follow-up queries are capped at two and deduplicated against queries already issued.

\paragraph{Deep Reader rules (English rendering).}
\begin{quote}\small
You are in the \textbf{deep-reading stage}: the locating stage has found the relevant materials (see the material summary in the task); your job is to get every detail. Content transcribed verbatim in the materials (table rows, text copied from figures) may be trusted directly; stop when it suffices. \textbf{Visual attributes} (counts, colours, positions, expressions, chart values whose captions were cropped) \textbf{must be looked at}: pass the figure's \texttt{fragment\_id} (one or several) and one explicit question to \texttt{read\_media} and trust what it sees; caption text is only a summary and is not evidence. Use \texttt{fragment\_context} for the text around a hit and \texttt{query\_table} for exact table values. \textbf{You have no search tool}: the materials are those given in the task, and your whole job is to read them deeply; read before you conclude, and never assert ``not present'' from a caption. The descriptions under ``related figures/tables'' are truncated and the answer may be in the cut part; read the full description with \texttt{fragment\_context} first. When the question names a page, select figures by their page label and read all figures on that page, aggregating per-figure results yourself. If \texttt{read\_media} answers ``not in the image'', try another candidate; only after reading may you conclude that the question is unanswerable.

Before reading a figure whose identity is uncertain (number, topic), use \texttt{media\_context} to see the caption and body text on its page, or \texttt{find\_figure} by number. For \textbf{counting questions} with a \texttt{list\_media} list in the materials, pass every \texttt{fragment\_id} in the list (at most 40) to \texttt{read\_media} in one call phrased as a yes/no test and count the per-figure results; never estimate from a few.
\end{quote}

\paragraph{Single-agent rules for P0.}
P0 uses the two-stage narrative of an earlier single-agent version (``your tool rounds are a scarce budget; spend the first 2--3 on locating, the rest on deep reading; when results largely repeat what you have, locating is done and you must switch to reading rather than re-searching''), plus the enumeration, figure-number, two-sided and truncated-caption rules above. Its tool set is the union of both roles' tools and its round cap is 8. Judge, pool, synthesizer and checker are identical to M5.

\paragraph{Benchmark-specific rule paragraph.}
M3DocVQA (all questions answerable): ``The answer to this question is certainly in the knowledge base: when evidence is partial give the most likely answer the materials support; for yes/no questions answer yes or no.'' MultiHop-RAG: ``This benchmark has insufficient-evidence questions: when the materials truly cannot answer, stop with \emph{Insufficient information} rather than forcing an answer; when they can, give the shortest entity or value; yes/no questions are answered Yes or No; for questions about order, earliest or latest, compare the Published dates at the top of each article.'' MMLongBench-Doc uses no extra paragraph.

\paragraph{Guardrail appended to every question (M3DocVQA version, verbatim).}
\begin{quote}\small\ttfamily
Answer requirements: Respond in English. Base your answer on the documents in this knowledge base; analyze as usual. Every question in this benchmark IS answerable from the knowledge base --- never reply `Not answerable', `not found' or `no information': if the evidence is partial, commit to the single most likely answer supported by what you found; for yes/no questions answer yes or no. End your reply with a single line `FINAL ANSWER: <answer>' where <answer> is the shortest exact form copied from the document: a single value/name/phrase with no extra or explanatory words; keep units, \% signs and file extensions exactly as written; colors as color names; integers as digits; if the question asks for two or more things, a JSON list with one element per thing in the order asked. If your analysis states the value, FINAL ANSWER must be that value.
\end{quote}
The MMLongBench-Doc version drops the ``every question is answerable'' sentence; the MultiHop-RAG version replaces it with the article-header, date-comparison, Yes/No and `Insufficient information' conventions.

\paragraph{Format checker (M3DocVQA version, verbatim).}
\begin{quote}\small\ttfamily
You check the last line `FINAL ANSWER: ...' of an answer written for an open-domain document QA benchmark in which EVERY question has an answer. You get the question and the full answer text. Rewrite only the FINAL ANSWER value, never add information that is not in the answer text. Rules: 1. Never output a refusal (`Not answerable', `None', `N/A', `not found', `no information' or any sentence saying the answer is missing). If the current FINAL ANSWER is a refusal but the text names a candidate value/entity for what the question asks, FINAL ANSWER must be that candidate. If the text gives no candidate at all, keep the FINAL ANSWER as it is. 1b. For yes/no questions the value is exactly `yes' or `no' as concluded by the text. 2. Shortest exact form, copied VERBATIM as a substring of the text: no explanatory or extra words; never start with an article; never paraphrase; if the text gives both a full phrase and its abbreviation, use the full phrase unless the question asks for the abbreviation; keep units, \% signs and file extensions exactly; keep original capitalization; colors as plain color names; integer counts as digits only; ranges as written. 2b. If the answer text has no `FINAL ANSWER' line at all, derive it from the text's own conclusion only. 3. If the question asks for two or more things, output a JSON list with one element per thing, in the order asked. Output JSON only: \{"final": "<value>"\} or \{"final": ["a", "b"]\}.
\end{quote}
The MMLongBench-Doc checker instead requires exactly `Not answerable' whenever the text says the premise does not match the document or the information is absent (``a wrong answer is worse than `Not answerable'\,''); the MultiHop-RAG checker maps every refusal to exactly `Insufficient information' and yes/no to `Yes'/`No'. The checker sees only the question and the answer text, never the materials, so it can only shorten or normalise, not add.

\paragraph{Negative results on prompts.}
Two alternatives were tested on the 35B model and rejected: the official M3DocRAG \texttt{text\_rag} template in place of our synthesizer ($-3.3$ F1 on mid300, $-1.4$ on the full set inside the multi-round pipeline), and a ``positive phrasing'' rewrite that removed every prohibition ($-4.6$ on mid300). On the 7B model the official template was the best of the three, so prompt choice is model-dependent and we did not reuse the small model's prompt on the large one. Two MultiHop-RAG variants that forced a Yes/No conclusion when two articles agree were $-12.9$ on a 31-question smoke set and were reverted before the full run.

\section{Full M3DocVQA tables}
\label{app:m3d}

\begin{table*}[t]
\centering\small
\resizebox{\textwidth}{!}{%
\begin{tabular}{@{}llrrrrrrrr@{}}
\toprule
Config & Description & R@1/3/8 (1st search) & R@all & Doc found & Ans.\ visible & EM & F1 & vs.\ ref. & Time p50 \\
\midrule
S0 & closed book, Qwen2-VL-7B & -- & -- & -- & -- & 17.2 & 22.0 & & \\
S0 & closed book, Qwen3.6 & -- & -- & -- & -- & 26.8 & 32.2 & & \\
S0 & closed book, Flash & -- & -- & -- & -- & 35.6 & 43.0 & & \\
\midrule
S1 & dense only, one shot & 53.5/72.5/80.3 & 82.6 & 71.4 & 66.6 & 32.8 & 39.7 & & 4.3s \\
S3 & dense+BM25, one shot & 56.3/77.4/84.5 & 86.3 & 75.6 & 68.3 & 33.3 & 40.6 & 176:156 vs S1 & \\
S4v1 & dense+BM25+graph, one shot, our prompt & 56.3/75.0/82.0 & 84.5 & 73.3 & 66.4 & 33.3 & 40.2 & 174:165 vs S1 & \\
S4 & same index, official M3DocRAG template & 56.3/75.0/82.0 & 84.5 & 73.3 & 66.4 & 33.1 & 40.4 & 222:210 vs S1 & 2.3s \\
S4 & same, Qwen2-VL-7B & 56.3/75.0/82.0 & 84.5 & 73.3 & 66.4 & 31.7 & 38.0 & & 3.1s \\
\midrule
M4 & pipeline, 12K pool, no judge & 56.9/72.4/78.7 & 89.6 & 81.0 & 73.9 & 37.7 & 45.6 & & 40.7s \\
M5 & full pipeline & 57.2/72.7/78.8 & 91.6 & 84.6 & 77.2 & 39.5 & 47.5 & 206:155 vs M4 & 38.2s \\
M5OFF & M5 with official template & & & & & & 46.1 & & \\
M5VEC & M5 on dense-only index & 51.1/66.3/72.5 & 89.9 & 81.8 & 75.4 & 37.3 & 45.0 & 155:226 vs M5 & 41.1s \\
M5S3 & M5 without graph expansion & 57.3/74.9/80.7 & 92.2 & 85.5 & 76.5 & 38.9 & 47.0 & 180:201 vs M5 & 38s \\
P0 & single agent, all tools, 8 rounds & 56.9/72.2/78.4 & 92.3 & 86.0 & 77.8 & 38.0 & 45.6 & 157:215 vs M5 & 34.6s \\
P1 & M5 without Deep Reader & --/--/78.7 & 90.9 & 83.3 & 76.4 & 35.8 & 43.7 & 149:247 vs M5 & 23s \\
M5 & full pipeline, Flash & 57.1/72.1/78.2 & 91.3 & 84.0 & 75.5 & 44.6 & 53.5 & 315:163 vs M5 & 27.0s \\
M4 & pipeline, Flash & & & & & 41.7 & 50.8 & & 19.1s \\
\bottomrule
\end{tabular}}
\caption{All full-set M3DocVQA runs (2,441 questions). ``vs.\ ref.'' gives question flips won:lost against the named reference. Time is the median per question under the 4-way concurrent harness; serial latencies are in \S\ref{sec:efficiency}.}
\label{tab:m3d-all}
\end{table*}

\begin{table}[t]
\centering\small
\resizebox{\columnwidth}{!}{%
\begin{tabular}{@{}lrrrrr@{}}
\toprule
F1 by bucket & image & table & text & multi & single \\
\midrule
S0 closed, Qwen3.6 & 22.2 & 25.1 & 40.9 & 21.9 & 39.1 \\
S0 closed, Flash & 29.1 & 35.6 & 54.8 & 32.6 & 50.0 \\
S1 dense one shot & 24.6 & 28.4 & 54.2 & 28.5 & -- \\
S3 hybrid one shot & 25.6 & 30.7 & 54.8 & 31.1 & -- \\
S4 one shot (official) & 23.5 & 30.4 & 53.7 & 27.7 & 49.0 \\
S4 one shot, 7B (official) & 25.8 & 28.3 & 49.3 & 27.7 & 44.9 \\
M4 & 27.8 & 37.7 & 58.6 & 34.4 & 53.1 \\
M5 & 29.5 & 40.0 & 60.5 & 36.6 & 54.9 \\
M5S3 & 29.6 & 40.3 & 59.7 & 37.5 & 53.4 \\
P0 & 27.8 & 37.8 & 58.9 & 34.9 & 52.9 \\
P1 & 27.2 & 34.5 & 59.0 & 34.9 & 49.7 \\
M5, Flash & 35.9 & 51.3 & 64.7 & 46.2 & -- \\
\bottomrule
\end{tabular}}
\caption{M3DocVQA F1 by evidence modality and hop count. Bucket sizes: image 940, table 1,265, text 1,216 (a question can carry several evidence modalities), multi-hop 980, single-hop 1,461; -- = not computed for that run.}
\label{tab:m3d-buckets}
\end{table}

\paragraph{Pool cap.} Of the 2,441 questions, 811 exceeded the 30K-token pool and were trimmed; M5 gains over M4 on both groups (untrimmed 53.9$\rightarrow$56.0, trimmed 28.9$\rightarrow$30.6), so the cap itself is neutral and the value of the pool is that the reader sees everything up to 30K rather than the first 12K.

\paragraph{Noise.} Three full runs of the 4+4 configuration on mid300 give 51.1 / 50.7 / 50.9; two full-set runs of the earlier pipeline differ by 0.3 F1 with symmetric flips (179:184). The per-question paired F1 standard deviation is 0.344, giving a paired standard error of $\pm$2.0 on 300 questions and $\pm$0.7 on the full set.

\section{Round-cap sweep: usage}
\label{app:usage}

\begin{table}[t]
\centering\small
\resizebox{\columnwidth}{!}{%
\begin{tabular}{@{}lrrrrrr@{}}
\toprule
Cap $L{+}R$ & Loc.\ calls & Loc.\ at cap & Read calls & Zero reads & Read at cap & p50 / p90 \\
\midrule
0+0 & -- & -- & -- & -- & -- & 4s / 24s \\
1+0 & 2.11 & 100\% & 0 & 100\% & -- & 10s / 38s \\
1+1 & 2.09 & 100\% & 0.24 & 76\% & 70\% & 18s / 60s \\
2+2 & 2.45 & 70\% & 0.38 & 67\% & 40\% & 27s / 80s \\
3+3 & 3.08 & 56\% & 0.48 & 63\% & 22\% & 33s / 102s \\
1+4 & 2.15 & 100\% & 0.77 & 58\% & 19\% & 23s / 88s \\
3+4 & 3.11 & 57\% & 0.72 & 59\% & 20\% & 30s / 102s \\
4+3 & 3.55 & 52\% & 0.47 & 68\% & 23\% & 33s / 108s \\
4+1 & 3.49 & 49\% & 0.25 & 75\% & 66\% & 31s / 76s \\
4+4 (3 runs) & 3.45--3.58 & 51\% & 0.67--0.73 & 55--60\% & 17--18\% & 35s / 110s \\
5+5 & 3.98 & 45\% & 0.85 & 55\% & 16\% & 41s / 139s \\
6+6 & 4.45 & 44\% & 0.90 & 57\% & 15\% & 44s / 143s \\
\bottomrule
\end{tabular}}
\caption{Actual tool usage under each round cap (mid300, Qwen3.6, 4-way concurrent harness). Locator calls include the bridge judge's follow-up searches (at most 2). The Deep Reader makes no call on 55--76\% of questions under every cap; its call count depends only on its own cap, not on the Locator's.}
\label{tab:usage}
\end{table}

Cutting one role at a time (single-role lines in Table~\ref{tab:budget}): Locator 4$\rightarrow$3 costs 2.3 F1 and 4$\rightarrow$1 costs 2.1; Reader 4$\rightarrow$3 costs 0.6 and 4$\rightarrow$1 costs 3.3; cutting both to 3 costs 4.7. The costs are not additive: the two single cuts at 3 sum to 2.9, while cutting both together costs 4.7, so each role's fourth round matters more when the other role is also short of rounds.

\section{Index construction cost}
\label{app:cost}

\begin{table*}[t]
\centering\small
\resizebox{\textwidth}{!}{%
\begin{tabular}{@{}lrrrrrrrrr@{}}
\toprule
Corpus & Source & Docs & Markdown chars & Images & OCR pages & Embeddings & Pre-proc.\ tokens & Extraction tokens (text / mm) & Total tokens \\
\midrule
M3DocVQA & 1.02 GB PDF & 3,368 & 57.7M & 12,556 & 17,223 & 201,957 & 44.3M & 72.1M / 21.7M & 138M \\
MMLongBench-Doc & 666 MB PDF & 135 & 11.8M & 7,189 & 3,657 & 34,234 & 8.4M & 6.7M / 5.2M & 20M \\
MultiHop-RAG & 5 MB Markdown & 609 & 5.2M & 0 & 0 & 7,608 & 0 & 3.1M / 0 & 3.1M \\
\bottomrule
\end{tabular}}
\caption{One-time index construction cost by corpus, from the system's usage ledger. Pre-processing covers file classification and heading-tree extraction; extraction covers knowledge-point extraction (text) and image/table description (multimodal, ``mm''). Relation typing runs on recalled pairs only and is included in the text figure. Wall clock: M3DocVQA about 12 hours at the highest concurrency preset; MMLongBench-Doc a median of 120\,s per document (6.6 hours in total at the normal preset); MultiHop-RAG 62 minutes.}
\label{tab:cost}
\end{table*}

The resulting M3DocVQA graph has 332K knowledge-point nodes and 299K edges, 133K of them typed. Tokens per Markdown character are 2.4 (M3DocVQA), 1.7 (MMLongBench-Doc) and 0.6 (MultiHop-RAG): PDF corpora with figures cost three to four times more to index than clean text, and pre-processing alone is 30--41\% of the PDF cost.

\section{Transfer benchmark details}
\label{app:transfer}

\begin{table}[t]
\centering\small
\resizebox{\columnwidth}{!}{%
\begin{tabular}{@{}lrrrrr@{}}
\toprule
MMLongBench-Doc (judge acc.) & n & S4 & 1+1 & M5 & P1 \\
\midrule
Single-page & 499 & 25 & 35 & 43 & 26 \\
Cross-page & 360 & 11 & 19 & 25 & 18 \\
Unanswerable & 223 & 95 & 92 & 92 & 93 \\
\midrule
Format Int & 299 & 16 & 22 & 29 & 21 \\
Format Str & 250 & 24 & 35 & 43 & 26 \\
Format Float & 159 & 25 & 36 & 45 & 30 \\
Format List & 151 & 11 & 20 & 26 & 13 \\
\midrule
Source Text & 298 & 20 & 29 & 33 & 26 \\
Source Table & 217 & 17 & 34 & 40 & 23 \\
Source Chart & 178 & 15 & 24 & 31 & 17 \\
Source Figure & 301 & 14 & 23 & 30 & 18 \\
Source Layout & 118 & 15 & 19 & 30 & 16 \\
\bottomrule
\end{tabular}}
\caption{MMLongBench-Doc by position, answer format and evidence source (the benchmark's own bucket protocol, in which a question with several sources counts in each). Earlier runs of the previous pipeline version on the same questions scored 46.2 / 44.3 / 45.4 (Qwen3.6) and 49.0 / 48.7 (Flash).}
\label{tab:mmlb-buckets}
\end{table}

\begin{table}[t]
\centering\small
\resizebox{\columnwidth}{!}{%
\begin{tabular}{@{}lrrrrr@{}}
\toprule
MultiHop-RAG, M5 & n & R@1 & R@3 & R@8 & R@all \\
\midrule
Inference & 816 & 20.5 & 38.8 & 58.3 & 73.0 \\
Comparison & 856 & 32.4 & 47.6 & 60.4 & 91.6 \\
Temporal & 583 & 30.4 & 50.3 & 65.6 & 88.9 \\
All non-null & 2,255 & 27.6 & 45.1 & 61.0 & 84.2 \\
\bottomrule
\end{tabular}}
\caption{MultiHop-RAG document-level retrieval for the full pipeline: recall of gold evidence articles in the first search (R@$k$) and in the union of all searches (R@all). The official benchmark reports chunk-level Hits@$k$ (best published: 0.747 at $k{=}10$ with a reranker), so the two are not directly comparable. Inference questions have low first-search recall because they typically need three or four articles.}
\label{tab:mhr-retrieval}
\end{table}

\paragraph{Refusals.} On MultiHop-RAG the format checker maps every refusal to `Insufficient information'. The pipeline predicted it on 705 / 612 / 515 / 532 questions (S4 / 1+1 / M5 / P1), of which 409 / 318 / 220 / 240 were on answerable questions. Prompt variants that reduced refusals further did so by forcing conclusions and lost more on comparison questions than they gained (Appendix~\ref{app:prompts}).

\section{Worked example}
\label{app:example}
Question (M3DocVQA; supporting documents \emph{1979 Trenton Twin Indy} and \emph{List of Indianapolis 500 winners}; gold ``Rick Mears''): \emph{Which driver with a time/retired of +1 lap in the Trenton Twin Indy of 1979 has won the Indy 500 the most times?} The race page's prose describes the two races and names the leaders; the ``Time/Retired'' column exists only in the results table, which is stored as an image fragment. All traces below are taken verbatim from the run records.

\paragraph{S4 one-shot (2.1\,s).} One hybrid search with the question retrieves the race page (plus three unrelated hits). The synthesizer answers from the prose: ``Al Unser''. Wrong.

\paragraph{P1 iterative retrieval, no reader (39.6\,s, 6 searches).} Locator: ``1979 Trenton Twin Indy results +1 lap retired'' $\to$ ``\ldots results table +1 lap finishers'' $\to$ ``\ldots full results +1 lap driver'' $\to$ ``\,`1979 Trenton Twin Indy' `+1 lap' driver'', all landing on the race page. Bridge judge: \texttt{need\_more=true}, reason ``the materials identify the drivers who finished +1 lap (Bobby Unser, Gordon Johncock, Al Unser, Johnny Rutherford) but do not contain their Indy 500 win counts''; the candidate list is inferred from the prose and is wrong. Follow-ups ``Bobby Unser Indy 500 wins'' and ``Gordon Johncock Indy 500 wins'' retrieve the \emph{List of Indianapolis 500 winners}. Pool: 25K characters, nothing evicted. Synthesizer: a long argument about which drivers were probably a lap down, ending in ``Al Unser''. Wrong.

\paragraph{M5 full pipeline (46.0\,s, 6 searches, 2 read calls).} The same four Locator searches and the same (wrong) judge candidates; the follow-up search retrieves the winners list. Deep Reader: \texttt{read\_media} on three table images, then on one more; the Race~6 results table returns the ``+1 Lap'' rows in positions 4--7: Wally Dallenbach, Rick Mears, Tom Sneva, Danny Ongais. Pool: 26.6K characters, nothing evicted. Synthesizer: identifies the four drivers from the read table, looks up their Indy~500 wins in the winners list already in the pool, and answers ``Rick Mears'' (four wins). Correct.

\paragraph{P0 single agent, all tools (31.9\,s, 5 searches, 2 read calls).} Search, then \texttt{read\_media} on the results table immediately, then a search for the four names' Indy~500 wins, a second read, two more searches; bridge judge follow-ups for Dallenbach and Sneva; answer ``Rick Mears''. Correct, and faster than M5 on this question. This is the honest counterpart of Table~\ref{tab:role}: a single agent \emph{can} read early; on average it reads half as often as the Deep Reader and never on 74\% of questions.

\section{Error attribution sample}
\label{app:errors}
From the M5 run, 414 questions have their supporting document as the first search hit yet F1 $<$ 0.5. A random 100 were classified into seven classes by DeepSeek-V4-Flash from the question, gold, prediction and trace, then re-checked independently by a second LLM reviewer (Claude) with the same material. The two passes agreed on 88 questions. All 12 disagreements were of one kind: 11 image questions on which the system refused (``not answerable'', ``not found'') were labelled C (image not read) by the first pass and F (refusal) by the second; we keep C, because the failure is that the figure was never opened, and reserve F for refusals whose answer was present in the retrieved \emph{text}. One synonym case (``spear gun'' for ``harpoon'') was moved from C to E. Final counts: C misread or unopened image 31; E correct answer at a granularity the metric penalises 23; A wrong field in the right document 19; F refusal with the gold in the text materials 13; B wrong table row or column 9; D multi-hop assembly 5; G other 0. By modality: text questions are dominated by A and E, table questions by B and A, image questions by C.

\end{document}